\documentclass{article}

    \PassOptionsToPackage{numbers, compress}{natbib}

\usepackage[final]{neurips_2024}

\usepackage[utf8]{inputenc} 
\usepackage[T1]{fontenc}    
\usepackage{hyperref}       
\usepackage{url}            
\usepackage{booktabs}       
\usepackage{amsfonts}       
\usepackage{nicefrac}       
\usepackage{microtype}      
\usepackage{xcolor}         

\usepackage{enumitem}
\usepackage{xspace}
\usepackage{xcolor}
\usepackage{wrapfig}
\usepackage{caption}
\usepackage{amssymb}
\usepackage{amsmath}
\usepackage{multirow}
\usepackage{adjustbox}
\usepackage{makecell}
\usepackage{subcaption}
\usepackage{algorithm}
\usepackage{algpseudocode}
\usepackage{colortbl}
\usepackage{mathtools}
\usepackage{setspace}
\usepackage{enumitem}
\usepackage{babel}
\usepackage{bbding}
\usepackage{utfsym}
\usepackage[flushleft]{threeparttable}

\usepackage{makecell}

\usepackage{thmtools}
\usepackage{thm-restate}

\newcommand{\baseline}{S2F\xspace}
\newcommand{\model}{S3F\xspace}
\newcommand{\modelmsa}{\model-MSA\xspace}

\newcommand{\cross}{\scalebox{0.75}{\usym{2613}}}

\definecolor{myred}{HTML}{e63946}
\definecolor{myblue}{HTML}{457b9d}
\newcommand{\first}[1]{\textcolor{myred}{#1}}

\setlist[enumerate]{leftmargin=7.5mm}

\usepackage{amsmath}
\usepackage{bm}

\def\eqref#1{equation~\ref{#1}}

\def\1{\bm{1}}

\def\ve{{\bm{e}}}
\def\vf{{\bm{f}}}

\def\vh{{\bm{h}}}

\def\vx{{\bm{x}}}

\def\mS{{\bm{S}}}

\def\mX{{\bm{X}}}

\DeclareMathAlphabet{\mathsfit}{\encodingdefault}{\sfdefault}{m}{sl}
\SetMathAlphabet{\mathsfit}{bold}{\encodingdefault}{\sfdefault}{bx}{n}

\def\gN{{\mathcal{N}}}

\newcommand{\R}{\mathbb{R}}

\let\oldAA\AA
\renewcommand{\AA}{\text{\normalfont\oldAA}}
\usepackage{cancel}

\title{Multi-Scale Representation Learning for\\Protein Fitness Prediction}

\author{
\textbf{Zuobai Zhang}\textsuperscript{\rm 1,2,*}\; 
\textbf{Pascal Notin}\textsuperscript{\rm 3,*}\; 
\textbf{Yining Huang}\textsuperscript{\rm 3}\; 
\textbf{Aur\'{e}lie Lozano}\textsuperscript{\rm 5}\\
\textbf{Vijil Chenthamarakshan}\textsuperscript{\rm 5}\;
\textbf{Debora Marks}\textsuperscript{\rm 3,4}\; 
\textbf{Payel Das}\textsuperscript{\rm 5,$\dagger$}\; 
\textbf{Jian Tang}\textsuperscript{\rm 1,6,7,$\dagger$}\\
\textsuperscript{\rm *}{\small equal contribution} \quad
\textsuperscript{\rm $\dagger$}{\small corresponding author} \\
\textsuperscript{\rm 1}Mila - Qu\'ebec AI Institute, \quad
\textsuperscript{\rm 2}Universit\'e de Montr\'eal, \quad 
\textsuperscript{\rm 3}Harvard Medical School, \\
\textsuperscript{\rm 4}Broad Institute, \quad
\textsuperscript{\rm 5}IBM Research, \quad
\textsuperscript{\rm 6}HEC Montr\'eal, \quad
\textsuperscript{\rm 7}CIFAR AI Chair \\
\texttt{zuobai.zhang@mila.quebec}, \quad \texttt{pascal\_notin@hms.harvard.edu}, \\
\texttt{daspa@us.ibm.com}, \quad \texttt{jian.tang@hec.ca} 
}

\begin{document}

\maketitle

\begin{abstract}
Designing novel functional proteins crucially depends on accurately modeling their fitness landscape. Given the limited availability of functional annotations from wet-lab experiments, previous methods have primarily relied on self-supervised models trained on vast, unlabeled protein sequence or structure datasets. While initial protein representation learning studies solely focused on either sequence or structural features, recent hybrid architectures have sought to merge these modalities to harness their respective strengths. However, these sequence-structure models have so far achieved only incremental improvements when compared to the leading sequence-only approaches, highlighting unresolved challenges effectively leveraging these modalities together. Moreover, the function of certain proteins is highly dependent on the granular aspects of their surface topology, which have been overlooked by prior models.
To address these limitations, we introduce the Sequence-Structure-Surface Fitness (\textbf{S3F}) model — a novel multimodal representation learning framework that integrates protein features across several scales. Our approach combines sequence representations from a protein language model with Geometric Vector Perceptron networks encoding protein backbone and detailed surface topology. The proposed method achieves state-of-the-art fitness prediction on the ProteinGym benchmark encompassing 217 substitution deep mutational scanning assays, and provides insights into the determinants of protein function.
Our code is at \url{https://github.com/DeepGraphLearning/S3F}.

\end{abstract}

\section{Introduction}

Proteins carry out a diverse range of functions in nature -- from catalyzing chemical reactions to supporting cellular structures, transporting molecules or transmitting signals. These functions are uniquely determined by their amino acid sequences and three-dimensional structures. The ability to design these sequences and structures presents significant opportunities to tackle critical challenges in sustainability, new material, and healthcare. 
This optimization process typically begins by learning the relationship between protein sequences or structures and their function, referred to as a \emph{fitness landscape}. 
This multivariate function describes how mutations impact protein fitness -- the more accurately we model these landscapes, the better we can engineer proteins with desired traits~\citep{Romero2009ExploringPF,Notin2024MachineLF}.

A significant challenge in modeling the fitness landscape is the scarcity of experimentally collected functional labels relative to the vastness of protein space~\citep{biswas2021low}. 
As a result, self-supervised approaches to protein representation learning have become crucial for predicting mutation effect~\citep{hopf2017mutation,meier2021language}.
While initial methods focused on learning a family-specific distribution over homologous protein sequences retrieved with a Multiple Sequence Alignment~\cite{hopf2017mutation,riesselman2018deep,frazer2021disease,laine2019gemme,shin2021protein}, subsequent methods have sought to learn general functional patterns across protein families, giving rise to `protein language models' or `family-agnostic models'~\citep{alley2019unified,rives2021biological,Madani2020ProGenLM}. Recently, hybrid models have achieved state-of-the-art fitness prediction performance by leveraging the relative strengths of both types of approaches~\citep{rao2021msa,notin2022tranception,notin2022trancepteve}.

Although sequence-based methods are effective in recapitulating certain aspects of protein structure~\cite{Morcos2011DirectcouplingAO,rao2021msa}, several protein functions and tasks crucially benefit from using a more granular representation of the protein structures and surfaces~\citep{Ingraham2019GenerativeMF,Baumann1989PolarityAA}.
To bridge this gap, recent studies have exploited advances in protein structure representation learning~\citep{jing2021equivariant,zhang2022protein,hsu2022learning}.
For instance, inverse folding models that learn a distribution over protein sequences conditioned on a protein backbone have shown enhanced performance in stability prediction~\citep{hsu2022learning, dauparas2022robust, yang2023masked, paul2023combining, Cagiada2024PredictingAP}. 
Recent efforts have also focused on integrating sequence-based and structure-based approaches ~\citep{su2024saprot,tan2023semantical}.
A prominent example, AlphaMissense, employs structural prediction losses to distill structural information into a hybrid model, highlighting the value from structural features~\citep{cheng2023accurate}. However, these hybrid sequence-structure methods have thus far only achieved modest improvements over leading sequence-based models, or have not made their model weights publicly available. Furthermore, current methodologies fall short in effectively modeling protein surfaces, which are essential for deciphering protein interactions and capturing broader structural details~\citep{gainza2020deciphering}.

\begin{figure}[t]
    \centering
    \includegraphics[width=\linewidth]{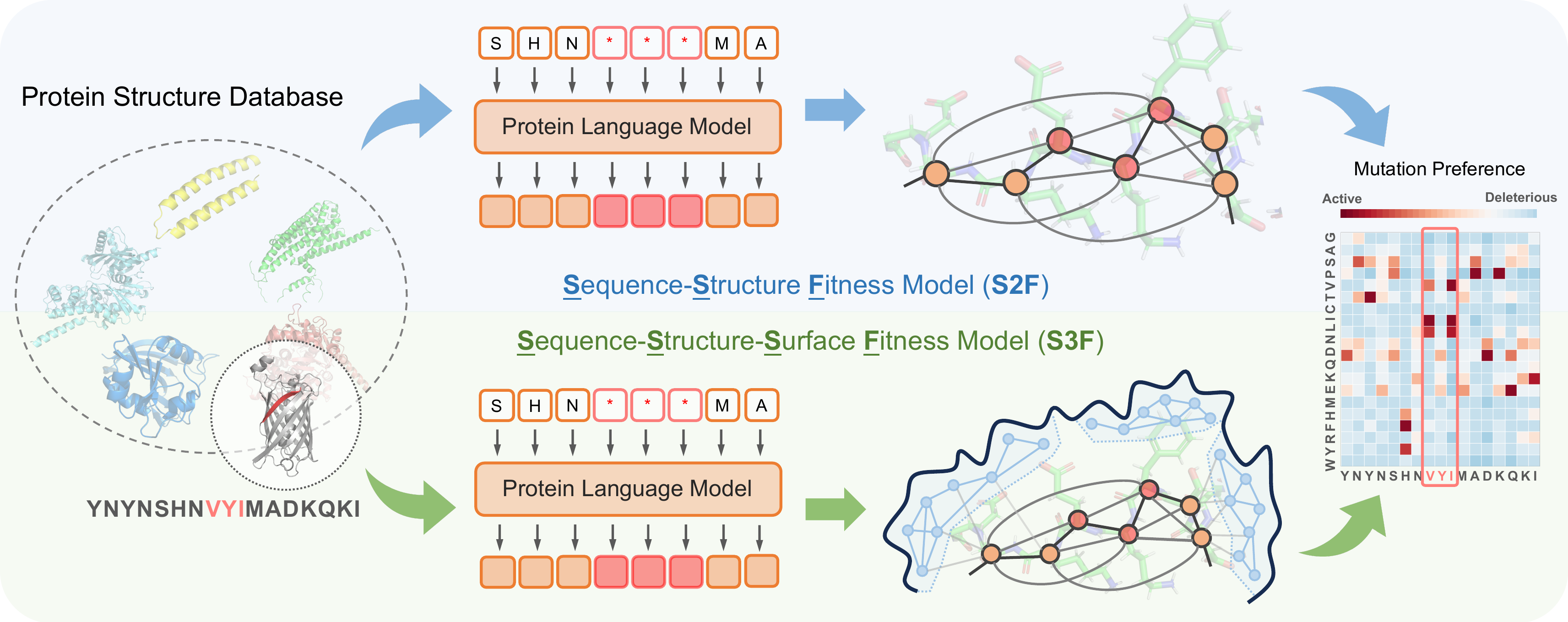}
    \caption{\textbf{Multi-scale Pre-training and Inference Frameworks for Protein Fitness Prediction}. During pre-training, protein sequences and structures are sampled from a database, with 15\% of residue types randomly masked. These sequences are fed into a protein language model, ESM-2-650M. Then, the output residue representations are used to initialize node features in our structure and surface encoders. Through message passing on structure and surface graphs, our methods, \textbf{\baseline} (blue) and \textbf{\model} (green), accurately predict the residue type distribution at each masked position. This distribution is subsequently used for mutation preferences in downstream fitness prediction tasks.}
    \label{fig:model}
\end{figure}

In this work, we introduce a multi-scale protein representation learning framework that integrates comprehensive levels of protein information for zero-shot protein fitness prediction (Fig.~\ref{fig:model}). We begin with a \textbf{S}equence-\textbf{S}tructure \textbf{F}itness Model (\textbf{\baseline}) by combining a protein language model with a structure-based encoder. \baseline utilizes the output of the protein language model as node features for a structure encoder, specifically a Geometric Vector Perceptron (GVP)~\citep{jing2021equivariant}, which enables message passing among spatially close neighborhoods. Building on this, we develop a \textbf{S}equence-\textbf{S}tructure-\textbf{S}urface \textbf{F}itness Model (\textbf{\model}), which enhances \baseline by adding a protein surface encoder that represents surfaces as point clouds and facilitates message passing between neighboring points. These multi-scale protein encoders are pre-trained using a residue type prediction loss on the CATH dataset~\citep{dawson2017cath}, enabling zero-shot prediction of mutation effects.

Our methods are rigorously evaluated using the comprehensive ProteinGym benchmark~\citep{notin2023proteingym}, which  includes 217 substitution deep mutational scanning assays and over 2.4 million mutated sequences across more than 200 diverse protein families. 
Our experimental results show that \baseline achieves competitive results with prior methods, while \model reaches state-of-the-art performance after incorporating surface features. When further augmented with alignment information, our method improves the current state-of-the-art by 8.5\% in terms of Spearman's rank correlation.
Additionally, our methods have substantially fewer trainable parameters compared to other baselines, reducing pre-training time to several days on commodity hardware. Being both lightweight and agnostic of the model used to obtain initial node embeddings, they can be readily adapted to augment forthcoming, more advanced protein language models.
To better understand the impact of multi-scale representation learning, we perform a breakdown analysis on different types of assays. Our results demonstrate the consistent improvements from multi-scale learning and show that incorporating structure and surface features can potentially correct biases in sequence-based methods, enhance accuracy in structure-related functions, and improve the ability to capture epistatic effects.
We summarize our contributions as follows:
\begin{itemize}[leftmargin=0.5cm]
    \item We develop a general and modular framework to learn \emph{multi-scale} protein representations (\S~\ref{sec:method});
    \item We introduce two instances of this framework -- \emph{\textbf{\baseline}} (\S~\ref{sec:s2f}) and \emph{\textbf{\model}} (\S~\ref{sec:s3f}), augmenting protein language model embeddings with structure and surface features for superior fitness prediction;
    \item We thoroughly evaluate our methods on the 217 assays from the ProteinGym benchmark, demonstrating their state-of-the-art performance and fast pre-training efficiency (\S~\ref{sec:overall});
    \item We perform a breakdown analysis for different types of assays to deep dive into the determinants of functions enabled by our multi-scale representation (\S~\ref{sec:breakdown}).
\end{itemize}

\section{Related Work}

\textbf{Protein Representation Learning.}
Previous research in protein representation learning has explored diverse modalities including sequences, multiple sequence alignments (MSAs), structures, and surfaces~\citep{lin2023evolutionary, rao2021msa, zhang2022protein, somnath2021multi}. Sequence-based methods treat protein sequences as a fundamental biological language, employing large-scale pre-training on billions of sequences to capture complex biological functions and evolutionary signals~\citep{tape2019, elnaggar2020prottrans, rives2021biological}. Alignment-based approaches, such as MSA Transformer~\citep{rao2021msa}, enhance representations by incorporating MSAs, improving the capture of evolutionary relationships. 
Recent advancements in structure prediction tools have shifted focus toward explicitly using protein structures for representation learning~\citep{gligorijevic2021structure, zhang2022protein, jing2021equivariant, hermosilla2020intrinsic}. These methods employ diverse self-supervised learning algorithms like contrastive learning, self-prediction, denoising, and masked structure token prediction to train structure encoders~\citep{zhang2022protein, chen2022structure, zhang2023physics, su2024saprot}.
Additionally, extracting features from protein surfaces has shown promise in uncovering critical chemical and geometric patterns important for biomolecular interactions~\citep{gainza2020deciphering, sverrisson2021fast,mallet2023atomsurf}. The growing availability of diverse protein data has also spurred the development of hybrid methods that combine multiple modalities for a holistic view~\citep{wang2022lm, zhang2023enhancing, somnath2021multi, lee2024pretraining,wu2023integration}.
Despite these advances, direct application of these models to zero-shot protein fitness prediction is still challenging and requires further design.

\textbf{Protein Fitness Prediction.}
Learning a fitness landscape has traditionally been approached as a discriminative supervised learning task, where models are trained to predict specific targets using labeled datasets~\citep{yang2019machine, gelman2021neural, dallago2021flip, notin2023proteinnpt}. 
Recently, unsupervised fitness predictors has shown promise in surpassing these traditional methods by overcoming the limitations and biases associated with sparse labels. 
These unsupervised models, often designed as protein language models, are trained on vast evolutionary datasets comprising millions of protein sequences, aiming to capture a general distribution across all proteins~\citep{nijkamp2023progen2,alley2019unified,notin2022tranception,rives2021biological,yang2024convolutions}. In contrast, alignment-based models focus on specific protein families, learning from multiple sequence alignments (MSAs) to capture nuanced distribution patterns~\citep{hopf2017mutation, riesselman2018deep, laine2019gemme, frazer2021disease, shin2021protein}. 
Additionally, hybrid sequence models integrate broad protein information with detailed family-specific data from alignments, enhancing the robustness of log-likelihood estimations and achieving top-tier performance~\citep{rao2021msa,notin2022trancepteve,truong2023poet}.

The incorporation of structural information into protein fitness prediction has marked a promising direction in the field, inspired by recent advances in structure representation learning. 
Based on the concept of protein language models, SaProt utilizes structure tokens from Foldseek~\citep{van2024fast} to generalize sequence-based methods to structural data~\citep{su2024saprot}. 
However, these structure-based methods only achieve limited improvement over the best sequence models.
The latest method, AlphaMissense, integrates structural prediction losses into a hybrid model, thereby enhancing predictive accuracy~\citep{cheng2023accurate}. 
Nevertheless, its high performance relies heavily on fine-tuning with weak supervision on human missense variants and it still lacks public accessibility to its model weights.
Moreover, to the best of our knowledge, no existing work has employed surface-based methods for fitness prediction. To fill this gap, this work aims to integrate multi-scale protein information for fitness prediction.

\section{Method}
\label{sec:method}

\subsection{Preliminary}

\textbf{Proteins.}
Proteins are macromolecules that form through the linkage of residues via peptide bonds. The three-dimensional (3D) structures of proteins are determined by the specific sequence of these residues. 
A protein with $n_r$ residues (amino acids) and $n_a$ atoms can be represented as a sequence-structure tuple $(\mS,\mX)$.
The sequence is denoted by $\mS=[s_1,s_2,\cdots,s_{n_r}]$, where $s_i\in\{1,...,20\}$ represents the type of the $i$-th residue. 
The structure is represented by $\mX=[\vx_1,\vx_2...,\vx_{n_a}]\in\R^{n_a\times 3}$, with $\vx_i$ specifying the Cartesian coordinates of the $i$-th atom.
For simplification, we only consider the alpha carbon atoms and ignore the side-chain variations induced by mutations.

\textbf{Protein Fitness Landscape.}
The ability of a protein to perform a specific function, often referred to as \emph{protein fitness}, is encoded by its sequence via spontaneous folding into structures. 
The effects of sequence mutations on protein function form a fitness landscape, which can be quantitatively measured through deep mutational scanning (DMS) experiments~\citep{fowler2014deep}. 
Modeling these landscapes is challenging due to the complicated relationship between sequences, structures and functions.

\textbf{Problem Definition.}
Unsupervised models that predict mutational effects are becoming fundamental tools in drug discovery, addressing the challenge of data scarcity. In this paper, we explore the task of zero-shot prediction of mutational effects using structural information. For simplicity, we focus solely on substitutions, leaving the study of insertions and deletions (indels) to future work.

Formally, for each DMS assay, we start with a wild-type protein \((\mathbf{S}^{\text{wt}}, \mathbf{X}^{\text{wt}})\) and generate a set of mutants by selecting specific mutation sites and randomly replacing the original residue type with a new one. For a mutant \((\mathbf{S}^{\text{mt}}, \mathbf{X}^{\text{mt}})\) with multiple mutations \(T\), the sequence changes such that \(s_t^{\text{mt}} \neq s_t^{\text{wt}}\) if \(t \in T\); otherwise, \(s_t^{\text{mt}} = s_t^{\text{wt}}\). We assume that the backbone structures remain unchanged post-mutation (\(\mathbf{X}^{\text{mt}} = \mathbf{X}^{\text{wt}}\)).
The objective is to develop an unsupervised model that can predict a score for each mutant to quantify the changes in fitness values relative to the wild-type.

\subsection{Protein Language Models for Mutational Effect Prediction}

Protein language models trained using the masked language modeling objective are designed to predict the likelihood of a residue's occurrence at a specific position within a protein, based on the surrounding context~\citep{rives2021biological,lin2023evolutionary}. 
As demonstrated in \citep{meier2021language}, this method can score sequence variations using the log odds ratio between mutant and wild-type proteins for given mutations $T$: 
\begin{equation}
\setlength{\abovedisplayskip}{5pt}
\setlength{\belowdisplayskip}{5pt}
\begin{matrix}
    \sum\nolimits_{t\in T} \log p(s_t=s_t^{mt}|\mS_{\backslash T}) - \log p(s_t=s_t^{wt}|\mS_{\backslash T}),
\end{matrix}
\end{equation}
where $\mS_{\backslash T}$ denotes the input sequence masked at each mutated position in $T$ and an additive model is assumed over multiple mutation sites. 
In the zero-shot setting, inference is performed directly on the sequence under evaluation using only forward passes of the model.

\subsection{Sequence-Structure Model for Fitness Prediction (\baseline)}
\label{sec:s2f}

Although these protein language models are effective in predicting mutational effects, they do not incorporate explicit structural information during pre-training, which is crucial for determining protein functions. Building on the ESM-2-650M model, we next propose \underline{\textbf{S}}equence-\underline{\textbf{S}}tructure \underline{\textbf{F}}itness Model (\textbf{\baseline}), which integrates structural data into the sequence-based predictive framework.

A major challenge in applying structure-based methods to fitness prediction is modeling how mutations impact protein structures. 
To bypass this problem, we adopt a simplified assumption: the backbone structures of proteins remain unchanged post-mutation. Additionally, we choose to omit side-chain information, which reveals residue types and could potentially lead to information leakage. 

\textbf{Geometric Message Passing.}
We build a radius graph for each protein with nodes representing alpha carbons. Two nodes are connected if their Euclidean distance is less than $10\AA$. We use Geometric Vector Perceptrons (GVP)~\citep{jing2021equivariant} to perform message passing across the graph. GVPs replace standard Multi-Layer Perceptrons (MLPs) in Graph Neural Networks, operating on scalar and geometric features that transform as vectors under spatial coordinate rotations.

We represent the hidden state of residue $i$ at the $l$-th layer by $\vh^{(l)}_i \in \R^d \times \R^{d' \times 3}$ with $d$-dim scalar features and $d'$-dim vector features. 
The initial node feature of residue $i$ is set using the protein language model embeddings, specifically, $\vh^{(0)}_i = (\text{ESM}(s_i|\mS_{\backslash T}), \mathbf{0})$. 
The edge features are given by $\ve_{(j,i)} = (\text{rbf}(\vx_j-\vx_i), \vx_j-\vx_i)$, where $\text{rbf}(\cdot)$ computes pairwise distance features with Radial Basis Function (RBF) kernels~\citep{rasmussen2010gaussian}. 
In the network, node and edge features are concatenated to facilitate message passing via the GVP module on the (scalar, vector) representations. Each message passing layer is followed by a feed-forward network. Formally,
\begin{align}
&\begin{matrix}\vh^{(l+0.5)}_{i} = \vh^{(l)}_{i} + \frac{1}{|\gN(i)|} \sum\nolimits_{j \in \gN(i)} \text{GVP}\left(\vh^{(l)}_{j}, \ve_{(j,i)}\right), \end{matrix}\\
&\begin{matrix}\vh^{(l+1)}_{i} = \vh^{(l+0.5)}_{i} + \text{GVP}\left(\vh^{(l+0.5)}_{i}\right),
\end{matrix}
\end{align}
where $\gN(i)$ represents the set of neighbors of node $i$. The GVP module ensures SE(3)-invariance for scalar features and SE(3)-equivariance for vector features. The scalar features at the last layer $\vh^{(L)}_i$ of each node $i$ are fed into a separate linear layer for predicting the residue type. Practically, we utilize $L=5$ layers of GVP, with $d'=16$ vector and $d=100$ scalar hidden representations.

\subsection{Sequence-Structure-Surface Model for Fitness Prediction (\model)}
\label{sec:s3f}

Besides protein sequences and structures, protein surfaces—defined by neighboring amino acids—are characterized by distinct patterns of geometric and chemical properties. For instance, within a folded protein, hydrophobic residues tend to cluster inside the core, while hydrophilic residues are exposed to water solvent on its surface. 
These patterns provide crucial insights into protein function and potential molecular interactions.
Now we integrate this aspect into our \baseline model to propose a new model called \underline{\textbf{S}}equence-\underline{\textbf{S}}tructure-\underline{\textbf{S}}urface \textbf{F}itness Model (\textbf{\model}).

\textbf{Surface Processing.}
We employ dMaSIF to generate the surface based on the backbone structure of each protein~\citep{sverrisson2021fast}. The surface is represented as a point cloud $\{\tilde{\vx}_1, \tilde{\vx}_2, ..., \tilde{\vx}_{n_s}\} \in \R^3$, consisting of $n_s$ (6K-20K) points. These points are sampled based on the levels of a smooth distance function defined over each atom. Each surface point $i$ is associated with geometric features $\tilde{\vf}_i$, specifically Gaussian curvatures~\citep{cao2019efficient} and Heat Kernel Signatures~\citep{sun2009concise}, providing a detailed characterization of the surface topology.
In the sequel, we will use the notations $\tilde{\cdot}$ with tilde for surface points to distinguish with those for nodes in structure graphs.

\textbf{Surface Feature Initialization.}
After generating protein surfaces, we build a mapping between the structure and surface graphs, projecting residue features onto the surface points. For each surface point $i$, we identify its 3 nearest neighboring residues, denoted as $\gN_{\text{surf-res}}(i)$, each initialized with its ESM feature. These features are concatenated with their Euclidean distances to point $i$ and processed through an MLP. The average features of these neighbors are then combined with the geometric features $\vf_i$. Formally, the scalar feature for surface node $i$ is initialized as follows:
\begin{equation}
    \tilde{\vh}_i^{(0)} = \text{MLP}\left(\vf_i, \begin{matrix}\frac{1}{3}\sum\nolimits_{j\in \gN_{\text{surf-res}(i)}} \text{MLP}(\vh_{j}^{(0)}, \|\tilde{\vx}_i-\vx_{j}\|_2)\end{matrix}\right),
\end{equation}
with the vector feature is initialized as the zero vector $\mathbf{0}$.

\textbf{Surface Message Passing.}
To perform message passing on protein surfaces, we construct a surface graph with a k-nearest neighbor graph, wherein each surface point is linked to its 16 nearest neighbors on the surface. For each edge $(j,i)$ in this graph, the edge feature $\tilde{\ve}{(j,i)}$ is initialized using RBF kernels and represented as $\tilde{\ve}{(j,i)} = (\text{rbf}(\tilde{\vx}_j-\tilde{\vx}_i), \tilde{\vx}_j-\tilde{\vx}_i)$.

Similar to \baseline, we employ GVP to perform message passing on the scalar and vector representations:
\begin{align}
&\begin{matrix}
\tilde{\vh}^{(l+0.5)}_{i} = \tilde{\vh}^{(l)}_{i} + \frac{1}{|\gN_{\text{surf}}(i)|} \sum\nolimits_{j \in \gN_{\text{surf}}(i)} \text{GVP}\left(\tilde{\vh}^{(l)}_{j}, \tilde{\ve}_{(j,i)}\right), \end{matrix}\\
&\begin{matrix}
\tilde{\vh}^{(l+1)}_{i} = \tilde{\vh}^{(l+0.5)}_{i} + \text{GVP}\left(\tilde{\vh}^{(l+0.5)}_{i}\right)\end{matrix},
\end{align}
where $\gN_{\text{surf}}(i)$ represents the neighbors of surface node $i$.
Similar with the approach on structure graphs, we utilize another five layers of GVP, with 16 vector and 100 scalar hidden representations.

\textbf{Residue Representation Aggregation.}
After performing message passing on both the structure and surface graphs, we combine the structure representations, $\vh^{(L)}$, with the surface representations, $\tilde{\vh}^{(L)}$. For each residue $i$, we identify the 20 nearest surface points, denoted as $\gN_{\text{res-surf}}(i)$, and compute their mean representations to enhance the residue's representation. 
Here we use more neighbors for mapping the surface points back to residues, as there are typically much more surface points than residues.
Specifically, we update each residue representation as follows:
\begin{equation}
    \vh_{i}^{(L)} \leftarrow \vh_{i}^{(L)} + \begin{matrix}\frac{1}{20}\sum_{j\in \gN_{\text{res-surf}}(i)} \tilde{\vh}_{j}^{(L)}\end{matrix},
\end{equation}
where $\leftarrow$ indicates that the left-hand side is updated with the value from the right-hand side. The updated representation $\vh_i^{(L)}$ is then input into a separate linear layer for predicting the residue type.

\subsection{Pre-Training and Inference}
\label{sec:pretrain}

Following the pre-training methodology in~\citep{devlin2018bert}, we select 15\% of residues at random for prediction. If the $i$-th residue is selected, we manipulate the $i$-th token by replacing it with: (1) the [MASK] token 80\% of the time, (2) a random residue type 10\% of the time, and (3) leaving the i-th token unchanged 10\% of the time. The final hidden state $\vh^{(L)}$ is then used to predict the original residue type using cross-entropy loss.
For \model, we avoid information leakage from surfaces by removing the top 20 closest surface points for each selected residue. 
During pre-training, the weights of the ESM-2-650M model are frozen, and only the GVP layers for structure and surface graphs are trainable. This strategy helps preserve sequence representations and improves pre-training efficiency.
Our models are pre-trained on a non-redundant subset of CATH v4.3.0 dataset (CC BY 4.0 license)~\citep{dawson2017cath}, which contains 30,948 experimental structures with less than 40\% sequence identity.
\baseline and \model are trained with batch sizes of 128 and 8, respectively, for 100 epochs on four A100 GPUs.
The pre-training time for \baseline and \model are 9 hours and 58 hours, respectively.

During inference, we adopt a strategy similar to that used in ESM-1v~\citep{meier2021language}. For each mutation, we mask the residue type at all mutation sites and remove the corresponding surface points, similar as the pre-training process. 
We use AlphaFold2 to predict the wild-type structures. 
To mitigate the influence of low-quality structures, for mutations on residues with a predicted Local Distance Difference Test (pLDDT) score below 70, we use the output scores from the baseline model, ESM-2-650M. For residues with pLDDT score no less than 70, we use the outputs from our \baseline or \model models.

\section{Experiment}
\label{sec:exp}

\subsection{Setup}
\label{sec:setup}

\textbf{Evaluation Dataset.}
To assess the performance of our method on zero-shot protein fitness prediction, we run experiments on the ProteinGym benchmark (MIT License)~\citep{notin2023proteingym}. 
This benchmark contains a comprehensive collection of Deep Mutational Scanning (DMS) assays, which covers a variety of functional properties such as thermostability, ligand binding, viral replication, and drug resistance. 
Specifically, we use 217 substitution assays that include both single and multiple mutations. 

\textbf{Metric.}
Given the complex, non-linear relationship between protein function and organism fitness~\citep{boucher2016quantifying}, we select the Spearman’s rank correlation coefficient as a primary metric for evaluating model prediction against experimental measurements. Besides, to ensure a comprehensive assessment, we include other metrics from the official ProteinGym benchmark: the Area Under the ROC Curve (AUC) and the Matthews Correlation Coefficient (MCC), which compare model scores with binarized experimental data. We also report the Normalized Discounted Cumulative Gains (NDCG) and Top 10\% Recall with the aim to identify the most functionally effective proteins.

\textbf{Baseline.}
We select a subset of baselines from the ProteinGym benchmark for comparison, categorizing them based on whether the model requires multiple sequence alignments (MSA) as input. 
In the category without MSAs, we include three protein language models—ProGen2 XL \citep{nijkamp2023progen2}, CARP-640M \citep{yang2024convolutions}, and ESM-2-650M \citep{lin2023evolutionary}; three inverse folding models—ProteinMPNN \citep{dauparas2022robust}, MIF \citep{yang2023masked}, and ESM-IF \citep{hsu2022learning}; and three sequence-structure hybrid models—MIF-ST \citep{yang2023masked}, ProtSSN \citep{tan2023semantical}, and SaProt \citep{su2024saprot}. 
For models utilizing alignments, we choose three family-specific models—DeepSequence \citep{riesselman2018deep}, EVE \citep{frazer2021disease}, and GEMME \citep{laine2019gemme}, as well as three hybrid models that combine family-agnostic and specific approaches, including MSA Transformer \citep{rao2021msa}, Tranception L with retrieval \citep{notin2022tranception}, and TranceptEVE \citep{notin2022trancepteve}. All baseline results are taken directly from the ProteinGym benchmark \citep{notin2023proteingym}.

We report the zero-shot performance of our methods, \baseline and \model. To benchmark against alignment-based models, we further enhance our models by ensembling them with EVE predictions through the summation of their z-scores, resulting in two variants named \baseline-MSA and \modelmsa.

\subsection{Benchmark Result}
\label{sec:overall}

\begin{table*}[t]
    \centering
    \caption{\textbf{Overall Results on ProteinGym.} 
    Models are categorized into two groups based on their reliance on MSA inputs. The types of input information and the number of trainable parameters are listed for each model. The best models within each category are highlighted in \textbf{\first{red}}.}
    \label{tab:proteingym}
    \begin{threeparttable}
    \begin{adjustbox}{max width=\linewidth}
        \begin{tabular}{ccccccccccccccccc}
            \toprule[2pt]
            \multirow{2}{*}{\bf{Model}} & \multicolumn{5}{c}{\bf{Benchmark Results}} && \multicolumn{5}{c}{\bf{Model Information}} \\ 
            \cmidrule{2-6}
            \cmidrule{8-12}
            & \bf Spearman & \bf AUC & \bf MCC & \bf NDCG & \bf Recall && \bf Seq. & \bf Struct. & \bf Surf. & \bf MSA & \bf \# Params. \\
            \midrule[1.5pt]
            ProGen2 XL &  0.391 & 0.717 & 0.306 & 0.767 & 0.199 && \multirow{3}{*}{\Large \checkmark} & \multirow{3}{*}{\Large \cross} & \multirow{3}{*}{\Large \cross} & \multirow{3}{*}{\Large \cross} & 6,400M\\
            CARP & 0.368 & 0.701 & 0.285 & 0.748 & 0.208 && &&& & 640M	\\
            ESM2 & 0.414 & 0.729 & 0.327 & 0.747 & 0.217 && &&& & 650M	\\
            \midrule
            ProteinMPNN &  0.258 & 0.639	 & 0.196	 & 0.713 & 0.186 && \multirow{3}{*}{\Large \cross} & \multirow{3}{*}{\Large \checkmark} & \multirow{3}{*}{\Large \cross} & \multirow{3}{*}{\Large \cross} & 2M \\
            MIF & 0.383 & 0.706 & 0.294 & 0.743 & 0.216 && &&& & 3M \\
            ESM-IF & 0.422 & 0.730 & 0.331 & 0.748 & 0.223 && &&& & 142M	\\
            \midrule
            MIF-ST & 0.383 & 0.717 & 0.310 & 0.765 & 0.226 && \multirow{3}{*}{\Large \checkmark} & \multirow{3}{*}{\Large \checkmark} & \multirow{3}{*}{\Large \cross} & \multirow{3}{*}{\Large \cross} & 643M	\\
            ProtSSN & 0.442 & 0.743 & 0.351 & 0.764 & 0.226 && &&& & 148M\\
            SaProt & 0.457 & 0.751 & 0.359 & 0.768 & {0.233} && &&& & 650M \\
            \midrule
            \bf \baseline & 0.454 & 0.749 & 0.359 & 0.762 & 0.227 && \multirow{2}{*}{\Large \checkmark} &\multirow{2}{*}{\Large \checkmark} & \cross & \multirow{2}{*}{\Large \cross} & 6M \\
            \bf \model & \bf \first{0.470} & \bf \first{0.757} & \bf \first{0.371}	& \bf \first{0.770} & \bf \first{0.234} && & & \checkmark & & 20M	\\
            \midrule[1.5pt]
            DeepSequence & 0.419 & 0.729 & 0.328 & 0.776	& 0.226 && \multirow{3}{*}{\Large \checkmark} &\multirow{3}{*}{\Large \cross}&\multirow{3}{*}{\Large \cross} & \multirow{3}{*}{\Large \checkmark} & 70M \\
            EVE & 0.439 & 0.741 & 0.342 & 0.783	 & 0.230 && &&& & 240M\\
            GEMME & 0.455 & 0.749 & 0.352 & 0.777 & 0.211 && &&& & <1M\\
            \midrule
            MSA Transformer & 0.434 & 0.738 & 0.340 & 0.779 & 0.224 && \multirow{3}{*}{\Large \checkmark} &\multirow{3}{*}{\Large \cross}&\multirow{3}{*}{\Large \cross} & \multirow{3}{*}{\Large \checkmark} & 100M	\\
            Tranception L & 0.434 & 0.739 & 0.341 & 0.779	 & 0.220 && &&& & 700M\\
            TranceptEVE L & 0.456 & 0.751 & 0.356 & 0.786 & 0.230 && &&& & 940M\\
            \midrule
            \bf \baseline-MSA & 0.487 & 0.767 & 0.381 & 0.790 & 0.240 && \multirow{2}{*}{\Large \checkmark} & \multirow{2}{*}{\Large \checkmark} & \cross & \multirow{2}{*}{\Large \checkmark} & 246M	\\
            \bf \modelmsa & \bf \first{0.496} & \bf \first{0.771} & \bf \first{0.387} & \bf \first{0.792} & \bf \first{0.244} &&  & & \checkmark & & 260M \\
            \bottomrule[2pt]
        \end{tabular}
    \end{adjustbox}
    \begin{tablenotes}
        \item[*] \scriptsize Since the numbers of trainable parameters for family-specific models depend on the length of protein sequences, we use the average \\ length (400 AAs) of all ProteinGym sequences for estimation.
    \end{tablenotes}
    \end{threeparttable}
    \vspace{-1em}
\end{table*}

\begin{figure}[t]
    \centering
    \includegraphics[width=\linewidth]{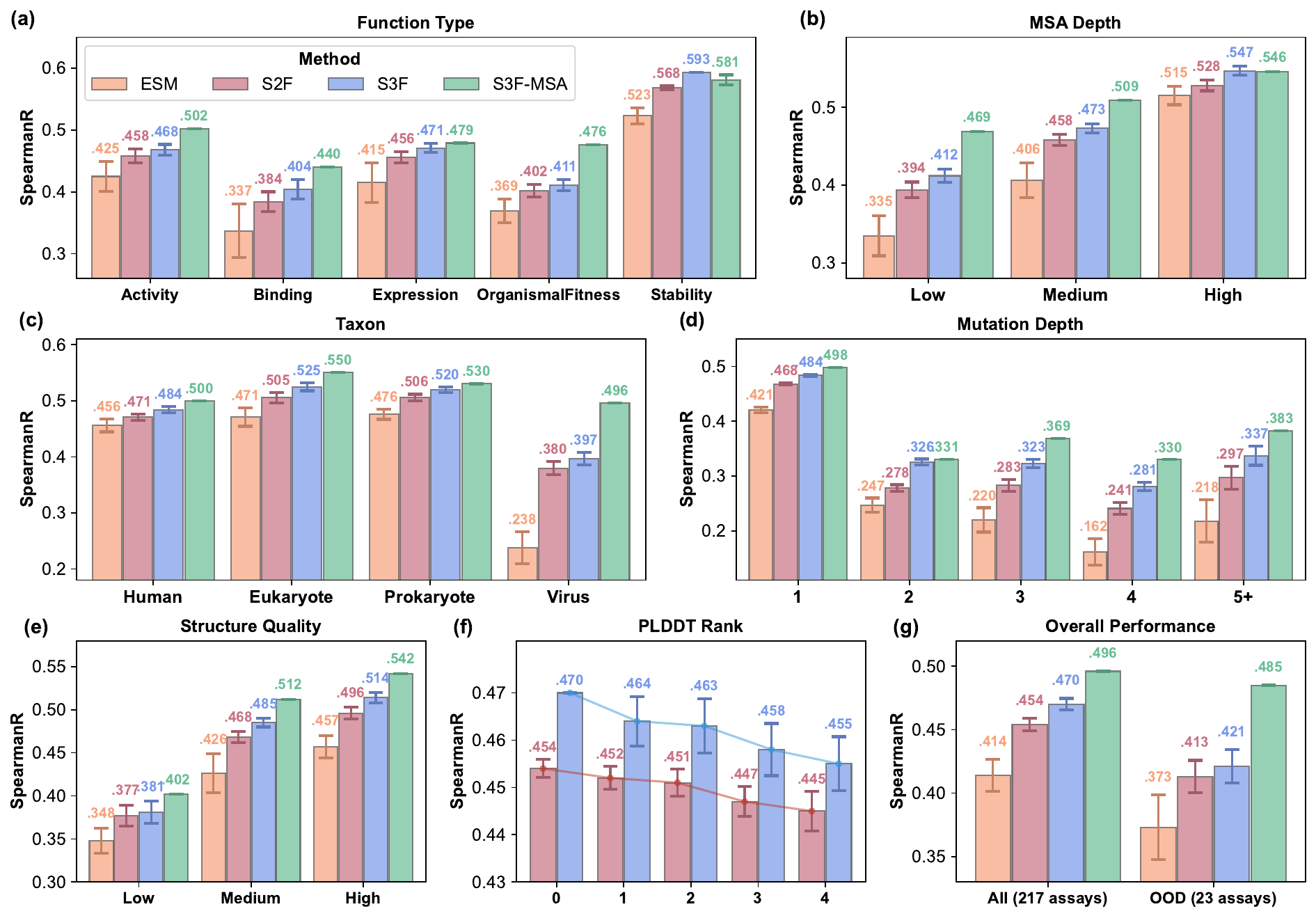}
    \caption{\textbf{Results of ESM-2-650M, \baseline, \model, and \modelmsa for Analyzing Contributions of Sequences, Structures, Surfaces, and Alignments.} \textbf{(a-d)} Breakdown performance (Spearman's rank correlation) on assays grouped by function type \textbf{(a)}, MSA depth \textbf{(b)}, taxon \textbf{(c)}, and mutation depth \textbf{(d)}. \textbf{(e-f)} Impact of protein structure quality on performance. \textbf{(e)} Breakdown performance on assays with low, medium, and high-quality structures. \textbf{(f)} Results using five groups of AlphaFold2-predicted structures ranked by pLDDT (0 for the highest pLDDT, 4 for the lowest pLDDT). \textbf{(g)} Results on all assays and out-of-distribution assays with low sequence similarity to the pre-training dataset.}
    \label{fig:breakdown}
    \vspace{-1em}
\end{figure}

We report the benchmark results and model information for all baselines in Table~\ref{tab:proteingym}. 
Among the methods that do not require MSAs, \baseline achieves competitive results, surpassing protein language models and inverse folding models, and only slightly lagging behind SaProt. When augmented with surface features, \model becomes the best model in this category, even outperforming the top alignment-based model, TranceptEVE, by a significant margin in terms of Spearman's rank correlation.
Despite relying on protein language models, our methods have substantially fewer trainable parameters compared to other baselines (20M in \model \emph{v.s.} 650M in SaProt). Hence, our methods finish pre-training within several days, whereas some large-scale baselines require months. 
Additionally, when enhanced with alignment information, \modelmsa improve the current state-of-the-art method, SaProt, by 8.5\% in terms of Spearman's rank correlation, further demonstrating the potential of our approach.
These results highlight both the effectiveness and the parameter efficiency of our proposed methods.

\subsection{Breakdown Analysis for Multi-Scale Learning}
\label{sec:breakdown}

From Table~\ref{tab:proteingym}, we observe the progressive improvements achieved by gradually incorporating various protein aspects into the model, as demonstrated by comparing the results of ESM2, \baseline, \model, and \modelmsa. To better understand the impact of this multi-scale representation learning, we conduct a breakdown analysis on different types of assays.
In Fig.~\ref{fig:breakdown}(a-d), we report the performance of these four methods on assays grouped by function types, MSA depths, taxon, and mutation depths. Our analysis reveals the following key points:\\[0.5em]
\textbf{1. Overall:} Consistent improvements are observed across all types of assays in different categories when structure, surface, and alignment information are added to the model. \\[0.3em]
\textbf{2. Function type:} Introducing structure and surface features significantly enhances performance on binding and stability assays. This aligns with our intuition that binding and stability are closely related to structural features, suggesting that structure-based methods have an advantage in identifying structure-disrupting mutations. This observation is consistent with previous studies \citep{paul2023combining}, while our methods also maintain competitive performance across other function types.\\[0.3em]
\textbf{3. MSA Depth:} As shown in Fig.~\ref{fig:breakdown}(b), protein language models perform poorly on assays with low MSA depths but excel on those with high MSA depths. This may be because proteins with low MSA depth are underrepresented in the ESM2 pre-training dataset, potentially reducing the diversity of specific families in protein language models. Introducing structure features partially mitigates this issue, and explicitly including family-specific training, like EVE, results in significant improvements.
\\[0.3em]
\textbf{4. Taxon:} 
Leveraging structure and surface-based features consistently improves the fitness prediction performance across taxa. Critically, when the underlying protein language model is poor for a given taxon (eg., ESM on viral proteins~\citep{notin2023proteingym}), incorporating structure and surface components provides inductive biases that help overcome these limitations. Developing taxa-specific models with higher quality node representations would likely results in further performance gains.\\[0.3em]
\textbf{5. Mutation Depth:} Most methods perform better on single mutations than on multiple mutations, likely due to our simplified additive assumption between mutations. As mutation depth increases, the performance gains from structure and surface encoding become more pronounced. This suggests that structure- and surface-based models are better at capturing epistatic effects.

In conclusion, multi-scale representation learning consistently improves performance. Incorporating structure and surface features can potentially correct biases in sequence-based methods, enhance accuracy in structure-related functions, and improve the ability to capture epistatic effects.

\subsection{Impact of Structure Quality}

To facilitate the usage of our methods on proteins without experimentally determined structures, we employ AlphaFold2 to generate structures for ProteinGym assays.
By default, five structures are generated for each assay, with the best one selected based on pLDDT scores.
Now we study how the quality of these predicted structures influences performance through two experiments.

First, we categorize the 217 assays into three groups based on the pLDDT of their predicted structures: 95 assays with pLDDT over 90 are classified as high-quality, 18 assays with pLDDT less than 70 as low-quality, and the remaining 104 assays as medium-quality. We report the average results for these categories in Fig.~\ref{fig:breakdown}(e).
The results indicate that performance for all four baselines positively correlates with structure quality, even for the sequence-based method, ESM. This may suggest our limited prior knowledge about these proteins with low-quality structures for all models, including AlphaFold2 and ESM2. Additionally, while improvements from structure and surface features are observed across all assays, those with high-quality structures benefit more significantly, especially for \model. This underscores the importance of accurate structures for fitness prediction.

Next, we analyze the impact of structure quality using the five AlphaFold2-predicted structures. We perform fitness predictions five times, each time using a different group of structures ranked by their pLDDT scores. Specifically, we test \baseline and \model using the top 1, 2, 3, 4, and 5 predicted structures for all assays, respectively. The results, plotted in Fig.~\ref{fig:breakdown}(f), show clear performance drops when lower-quality structures are used. This further highlights the reliance on high-quality structures.

\subsection{Generalization Ability to Unseen Protein Families}
\label{sec:generalization}

In contrast to large-scale models pre-trained on UniRef100 or the AlphaFold Database, our method employs a much smaller dataset, CATH, which contains only 31,000 structures after clustering. This raises the question of whether the benefits of structure- and surface-based methods can generalize well to unseen protein families, despite the limited scale of the pre-training data.
To address this, we select 23 out-of-distribution assays from ProteinGym, whose sequences have less than 30\% similarity to the pre-training dataset. We plot the average performance of the four methods on these assays in Fig.~\ref{fig:breakdown}(g). Although the absolute performance of all four methods decreases compared to their overall performance, we still observe consistent improvements from the structure- and surface-based methods, as evidenced by the performance gains of \baseline over ESM and \model over \baseline.
This proves the generalization ability of our methods to unseen protein families.

\subsection{Case Study: Epistatic effects in the IgG-binding domain of protein G}

\begin{figure}[t]
    \centering
    \includegraphics[width=\linewidth]{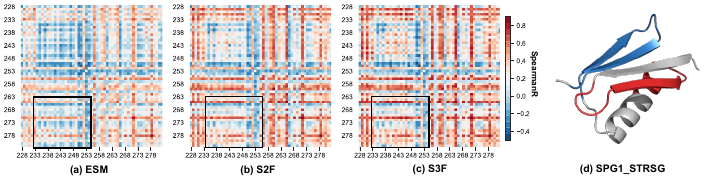}
    \caption{\textbf{Case Study on GB1.} \textbf{(a-c)} For each pair of mutation sites, we plot the Spearman's rank correlation between the experimental values and model-predicted scores for all 361 mutations: ESM (a), \baseline (b), and \model (c). The epistasis between residues 234-252 and residues 266-282 (in the black rectangle) are better captured by \baseline and \model. \textbf{(d)} Visualization of the predicted structure for GB1. Mutation regions 234-252 and 266-282 are highlighted in red and blue, respectively.}
    \label{fig:case_study}
\end{figure}



We illustrate the advantages of our method by focusing on the results from the high-throughput assay of the IgG-binding domain of protein G (GB1) introduced in~\citep{olson_comprehensive_2014}. The assay quantifies the effects of all single and double mutations, thereby providing detailed insights into the epistatic relationships between residue pairs in the domain. We report in Fig.~\ref{fig:case_study} the pair-specific Spearman's rank correlation between the experimental measurements and predictions from ESM, \baseline and \model, respectively. Leveraging structure and surface features leads to a superior ability in predicting mutation effects -- in particular the epistatic effects between residue pairs that are far in sequence space but close in the tertiary structure of GB1 (off-diagonal terms in Fig.~\ref{fig:case_study}(a-c) and colored residues in Fig.~\ref{fig:case_study}(d)).
\section{Conclusion and Limitations}
\label{sec:conclusion}

This work introduces \model, a novel multi-scale representation learning framework for protein fitness prediction that achieves state-of-the-art performance on the ProteinGym benchmark with a lightweight model. The breakdown analysis provides insights into how different protein modalities contribute to predicting various fitness landscapes. However, there are still several limitations in this work.
First, our models are pre-trained on relatively small set of experimentally determined protein structures and may further benefit from leveraging more diverse structure sets, such as the AlphaFold database.
Second, we make simplifying assumptions, such as ignoring side-chain information and assuming that backbone structures remain unchanged after mutations, which may limit the model's capacity. Third, our approach is limited to substitutions effects, and does not handle insertions nor deletions.

\section*{Acknowledgments}

The authors would like to thank Jiarui Lu and Chenqing Hua for their helpful discussions.

This project is supported by AIHN IBM-MILA partnership program, the Natural Sciences and Engineering Research Council (NSERC) Discovery Grant, the Canada CIFAR AI Chair Program, collaboration grants between Microsoft Research and Mila, Samsung Electronics Co., Ltd., Amazon Faculty Research Award, Tencent AI Lab Rhino-Bird Gift Fund, a NRC Collaborative R\&D Project (AI4D-CORE-06) as well as the IVADO Fundamental Research Project grant PRF-2019-3583139727.

P.N. was supported by a Chan Zuckerberg Initiative Award (Neurodegeneration Challenge Network, CZI2018-191853). D.M. holds a Ben Barres Early Career Award from the Chan Zuckerberg Initiative as part of the Neurodegeneration Challenge Network (CZI2018-191853) and is supported by a NIH Transformational Research Award (TR01 1R01CA260415).

\bibliography{reference}

\begin{thebibliography}{10}

\bibitem{Romero2009ExploringPF}
Philip~A. Romero and Frances~H. Arnold.
\newblock Exploring protein fitness landscapes by directed evolution.
\newblock {\em Nature Reviews Molecular Cell Biology}, 10:866--876, 2009.

\bibitem{Notin2024MachineLF}
Pascal Notin, Nathan~J. Rollins, Yarin Gal, Chris Sander, and Debora Marks.
\newblock Machine learning for functional protein design.
\newblock {\em Nature Biotechnology}, 42:216--228, 2024.

\bibitem{biswas2021low}
Surojit Biswas, Grigory Khimulya, Ethan~C Alley, Kevin~M Esvelt, and George~M Church.
\newblock Low-n protein engineering with data-efficient deep learning.
\newblock {\em Nature methods}, 18(4):389--396, 2021.

\bibitem{hopf2017mutation}
Thomas~A Hopf, John~B Ingraham, Frank~J Poelwijk, Charlotta~PI Sch{\"a}rfe, Michael Springer, Chris Sander, and Debora~S Marks.
\newblock Mutation effects predicted from sequence co-variation.
\newblock {\em Nature biotechnology}, 35(2):128--135, 2017.

\bibitem{meier2021language}
Joshua Meier, Roshan Rao, Robert Verkuil, Jason Liu, Tom Sercu, and Alex Rives.
\newblock Language models enable zero-shot prediction of the effects of mutations on protein function.
\newblock {\em Advances in neural information processing systems}, 34:29287--29303, 2021.

\bibitem{riesselman2018deep}
Adam~J Riesselman, John~B Ingraham, and Debora~S Marks.
\newblock Deep generative models of genetic variation capture the effects of mutations.
\newblock {\em Nature methods}, 15(10):816--822, 2018.

\bibitem{frazer2021disease}
Jonathan Frazer, Pascal Notin, Mafalda Dias, Aidan Gomez, Joseph~K Min, Kelly Brock, Yarin Gal, and Debora~S Marks.
\newblock Disease variant prediction with deep generative models of evolutionary data.
\newblock {\em Nature}, 599(7883):91--95, 2021.

\bibitem{laine2019gemme}
Elodie Laine, Yasaman Karami, and Alessandra Carbone.
\newblock Gemme: a simple and fast global epistatic model predicting mutational effects.
\newblock {\em Molecular biology and evolution}, 36(11):2604--2619, 2019.

\bibitem{shin2021protein}
Jung-Eun Shin, Adam~J Riesselman, Aaron~W Kollasch, Conor McMahon, Elana Simon, Chris Sander, Aashish Manglik, Andrew~C Kruse, and Debora~S Marks.
\newblock Protein design and variant prediction using autoregressive generative models.
\newblock {\em Nature communications}, 12(1):2403, 2021.

\bibitem{alley2019unified}
Ethan~C Alley, Grigory Khimulya, Surojit Biswas, Mohammed AlQuraishi, and George~M Church.
\newblock Unified rational protein engineering with sequence-based deep representation learning.
\newblock {\em Nature methods}, 16(12):1315--1322, 2019.

\bibitem{rives2021biological}
Alexander Rives, Joshua Meier, Tom Sercu, Siddharth Goyal, Zeming Lin, Jason Liu, Demi Guo, Myle Ott, C~Lawrence Zitnick, Jerry Ma, et~al.
\newblock Biological structure and function emerge from scaling unsupervised learning to 250 million protein sequences.
\newblock {\em Proceedings of the National Academy of Sciences}, 118(15), 2021.

\bibitem{Madani2020ProGenLM}
Ali Madani, Bryan McCann, Nikhil~Vijay Naik, Nitish~Shirish Keskar, Namrata Anand, Raphael~R. Eguchi, Po-Ssu Huang, and Richard Socher.
\newblock Progen: Language modeling for protein generation.
\newblock {\em bioRxiv}, 2020.

\bibitem{rao2021msa}
Roshan~M Rao, Jason Liu, Robert Verkuil, Joshua Meier, John Canny, Pieter Abbeel, Tom Sercu, and Alexander Rives.
\newblock Msa transformer.
\newblock In {\em Proceedings of the 38th International Conference on Machine Learning}, volume 139 of {\em Proceedings of Machine Learning Research}, pages 8844--8856. PMLR, 18--24 Jul 2021.

\bibitem{notin2022tranception}
Pascal Notin, Mafalda Dias, Jonathan Frazer, Javier~Marchena Hurtado, Aidan~N Gomez, Debora Marks, and Yarin Gal.
\newblock Tranception: protein fitness prediction with autoregressive transformers and inference-time retrieval.
\newblock In {\em International Conference on Machine Learning}, pages 16990--17017. PMLR, 2022.

\bibitem{notin2022trancepteve}
Pascal Notin, Lood Van~Niekerk, Aaron~W Kollasch, Daniel Ritter, Yarin Gal, and Debora~S Marks.
\newblock Trancepteve: Combining family-specific and family-agnostic models of protein sequences for improved fitness prediction.
\newblock {\em bioRxiv}, pages 2022--12, 2022.

\bibitem{Morcos2011DirectcouplingAO}
Faruck Morcos, Andrea Pagnani, Bryan Lunt, Arianna Bertolino, Debora~S. Marks, Chris Sander, Riccardo Zecchina, Jos{\'e}~Nelson Onuchic, Terence Hwa, and Martin Weigt.
\newblock Direct-coupling analysis of residue coevolution captures native contacts across many protein families.
\newblock {\em Proceedings of the National Academy of Sciences}, 108:E1293 -- E1301, 2011.

\bibitem{Ingraham2019GenerativeMF}
John Ingraham, Vikas~K. Garg, Regina Barzilay, and T.~Jaakkola.
\newblock Generative models for graph-based protein design.
\newblock In {\em DGS@ICLR}, 2019.

\bibitem{Baumann1989PolarityAA}
G~Baumann, Cornelius Fr{\"o}mmel, and Chris Sander.
\newblock Polarity as a criterion in protein design.
\newblock {\em Protein engineering}, 2 5:329--34, 1989.

\bibitem{jing2021equivariant}
Bowen Jing, Stephan Eismann, Pratham~N. Soni, and Ron~O. Dror.
\newblock Learning from protein structure with geometric vector perceptrons.
\newblock In {\em International Conference on Learning Representations}, 2021.

\bibitem{zhang2022protein}
Zuobai Zhang, Minghao Xu, Arian Jamasb, Vijil Chenthamarakshan, Aurelie Lozano, Payel Das, and Jian Tang.
\newblock Protein representation learning by geometric structure pretraining.
\newblock In {\em The Eleventh International Conference on Learning Representations}, 2023.

\bibitem{hsu2022learning}
Chloe Hsu, Robert Verkuil, Jason Liu, Zeming Lin, Brian Hie, Tom Sercu, Adam Lerer, and Alexander Rives.
\newblock Learning inverse folding from millions of predicted structures.
\newblock In {\em International conference on machine learning}, pages 8946--8970. PMLR, 2022.

\bibitem{dauparas2022robust}
Justas Dauparas, Ivan Anishchenko, Nathaniel Bennett, Hua Bai, Robert~J Ragotte, Lukas~F Milles, Basile~IM Wicky, Alexis Courbet, Rob~J de~Haas, Neville Bethel, et~al.
\newblock Robust deep learning--based protein sequence design using proteinmpnn.
\newblock {\em Science}, 378(6615):49--56, 2022.

\bibitem{yang2023masked}
Kevin~K Yang, Niccol{\`o} Zanichelli, and Hugh Yeh.
\newblock Masked inverse folding with sequence transfer for protein representation learning.
\newblock {\em Protein Engineering, Design and Selection}, 36:gzad015, 2023.

\bibitem{paul2023combining}
Steffanie Paul, Aaron Kollasch, Pascal Notin, and Debora Marks.
\newblock Combining structure and sequence for superior fitness prediction.
\newblock In {\em NeurIPS 2023 Generative AI and Biology (GenBio) Workshop}, 2023.

\bibitem{Cagiada2024PredictingAP}
Matteo Cagiada, Sergey Ovchinnikov, and Kresten Lindorff-Larsen.
\newblock Predicting absolute protein folding stability using generative models.
\newblock {\em bioRxiv}, 2024.

\bibitem{su2024saprot}
Jin Su, Chenchen Han, Yuyang Zhou, Junjie Shan, Xibin Zhou, and Fajie Yuan.
\newblock Saprot: Protein language modeling with structure-aware vocabulary.
\newblock In {\em The Twelfth International Conference on Learning Representations}, 2024.

\bibitem{tan2023semantical}
Yang Tan, Bingxin Zhou, Lirong Zheng, Guisheng Fan, and Liang Hong.
\newblock Semantical and topological protein encoding toward enhanced bioactivity and thermostability.
\newblock {\em bioRxiv}, pages 2023--12, 2023.

\bibitem{cheng2023accurate}
Jun Cheng, Guido Novati, Joshua Pan, Clare Bycroft, Akvil{\.e} {\v{Z}}emgulyt{\.e}, Taylor Applebaum, Alexander Pritzel, Lai~Hong Wong, Michal Zielinski, Tobias Sargeant, et~al.
\newblock Accurate proteome-wide missense variant effect prediction with alphamissense.
\newblock {\em Science}, 381(6664):eadg7492, 2023.

\bibitem{gainza2020deciphering}
Pablo Gainza, Freyr Sverrisson, Frederico Monti, Emanuele Rodola, D~Boscaini, Michael~M Bronstein, and Bruno~E Correia.
\newblock Deciphering interaction fingerprints from protein molecular surfaces using geometric deep learning.
\newblock {\em Nature Methods}, 17(2):184--192, 2020.

\bibitem{dawson2017cath}
Natalie~L Dawson, Tony~E Lewis, Sayoni Das, Jonathan~G Lees, David Lee, Paul Ashford, Christine~A Orengo, and Ian Sillitoe.
\newblock Cath: an expanded resource to predict protein function through structure and sequence.
\newblock {\em Nucleic acids research}, 45(D1):D289--D295, 2017.

\bibitem{notin2023proteingym}
Pascal Notin, Aaron Kollasch, Daniel Ritter, Lood Van~Niekerk, Steffanie Paul, Han Spinner, Nathan Rollins, Ada Shaw, Rose Orenbuch, Ruben Weitzman, et~al.
\newblock Proteingym: large-scale benchmarks for protein fitness prediction and design.
\newblock {\em Advances in Neural Information Processing Systems}, 36, 2023.

\bibitem{lin2023evolutionary}
Zeming Lin, Halil Akin, Roshan Rao, Brian Hie, Zhongkai Zhu, Wenting Lu, Nikita Smetanin, Robert Verkuil, Ori Kabeli, Yaniv Shmueli, et~al.
\newblock Evolutionary-scale prediction of atomic-level protein structure with a language model.
\newblock {\em Science}, 379(6637):1123--1130, 2023.

\bibitem{somnath2021multi}
Vignesh~Ram Somnath, Charlotte Bunne, and Andreas Krause.
\newblock Multi-scale representation learning on proteins.
\newblock {\em Advances in Neural Information Processing Systems}, 34:25244--25255, 2021.

\bibitem{tape2019}
Roshan Rao, Nicholas Bhattacharya, Neil Thomas, Yan Duan, Xi~Chen, John Canny, Pieter Abbeel, and Yun~S Song.
\newblock Evaluating protein transfer learning with tape.
\newblock In {\em Advances in Neural Information Processing Systems}, 2019.

\bibitem{elnaggar2020prottrans}
Ahmed Elnaggar, Michael Heinzinger, Christian Dallago, Ghalia Rehawi, Wang Yu, Llion Jones, Tom Gibbs, Tamas Feher, Christoph Angerer, Martin Steinegger, Debsindhu Bhowmik, and Burkhard Rost.
\newblock Prottrans: Towards cracking the language of lifes code through self-supervised deep learning and high performance computing.
\newblock {\em IEEE Transactions on Pattern Analysis and Machine Intelligence}, pages 1--1, 2021.

\bibitem{gligorijevic2021structure}
Vladimir Gligorijevi{\'c}, P~Douglas Renfrew, Tomasz Kosciolek, Julia~Koehler Leman, Daniel Berenberg, Tommi Vatanen, Chris Chandler, Bryn~C Taylor, Ian~M Fisk, Hera Vlamakis, et~al.
\newblock Structure-based protein function prediction using graph convolutional networks.
\newblock {\em Nature communications}, 12(1):1--14, 2021.

\bibitem{hermosilla2020intrinsic}
Pedro Hermosilla, Marco Sch{\"a}fer, Mat{\v{e}}j Lang, Gloria Fackelmann, Pere~Pau V{\'a}zquez, Barbora Kozl{\'\i}kov{\'a}, Michael Krone, Tobias Ritschel, and Timo Ropinski.
\newblock Intrinsic-extrinsic convolution and pooling for learning on 3d protein structures.
\newblock {\em International Conference on Learning Representations}, 2021.

\bibitem{chen2022structure}
Can~(Sam) Chen, Jingbo Zhou, Fan Wang, Xue Liu, and Dejing Dou.
\newblock Structure-aware protein self-supervised learning.
\newblock {\em Bioinformatics}, 39, 2022.

\bibitem{zhang2023physics}
Zuobai Zhang, Minghao Xu, Aurelie Lozano, Vijil Chenthamarakshan, Payel Das, and Jian Tang.
\newblock Pre-training protein encoder via siamese sequence-structure diffusion trajectory prediction.
\newblock In {\em Thirty-seventh Conference on Neural Information Processing Systems}, 2023.

\bibitem{sverrisson2021fast}
Freyr Sverrisson, Jean Feydy, Bruno~E Correia, and Michael~M Bronstein.
\newblock Fast end-to-end learning on protein surfaces.
\newblock In {\em Proceedings of the IEEE/CVF Conference on Computer Vision and Pattern Recognition}, pages 15272--15281, 2021.

\bibitem{mallet2023atomsurf}
Vincent Mallet, Souhaib Attaiki, and Maks Ovsjanikov.
\newblock Atomsurf: Surface representation for learning on protein structures.
\newblock {\em arXiv preprint arXiv:2309.16519}, 2023.

\bibitem{wang2022lm}
Zichen Wang, Steven~A Combs, Ryan Brand, Miguel~Romero Calvo, Panpan Xu, George Price, Nataliya Golovach, Emmanuel~O Salawu, Colby~J Wise, Sri~Priya Ponnapalli, et~al.
\newblock Lm-gvp: an extensible sequence and structure informed deep learning framework for protein property prediction.
\newblock {\em Scientific reports}, 12(1):6832, 2022.

\bibitem{zhang2023enhancing}
Zuobai Zhang, Chuanrui Wang, Minghao Xu, Vijil Chenthamarakshan, Aurelie Lozano, Payel Das, and Jian Tang.
\newblock A systematic study of joint representation learning on protein sequences and structures.
\newblock {\em arXiv preprint arXiv:2303.06275}, 2023.

\bibitem{lee2024pretraining}
Youhan Lee, Hasun Yu, Jaemyung Lee, and Jaehoon Kim.
\newblock Pre-training sequence, structure, and surface features for comprehensive protein representation learning.
\newblock In {\em The Twelfth International Conference on Learning Representations}, 2024.

\bibitem{wu2023integration}
Fang Wu, Lirong Wu, Dragomir Radev, Jinbo Xu, and Stan~Z Li.
\newblock Integration of pre-trained protein language models into geometric deep learning networks.
\newblock {\em Communications Biology}, 6(1):876, 2023.

\bibitem{yang2019machine}
Kevin~K Yang, Zachary Wu, and Frances~H Arnold.
\newblock Machine-learning-guided directed evolution for protein engineering.
\newblock {\em Nature methods}, 16(8):687--694, 2019.

\bibitem{gelman2021neural}
Sam Gelman, Sarah~A Fahlberg, Pete Heinzelman, Philip~A Romero, and Anthony Gitter.
\newblock Neural networks to learn protein sequence--function relationships from deep mutational scanning data.
\newblock {\em Proceedings of the National Academy of Sciences}, 118(48):e2104878118, 2021.

\bibitem{dallago2021flip}
Christian Dallago, Jody Mou, Kadina~E Johnston, Bruce~J Wittmann, Nicholas Bhattacharya, Samuel Goldman, Ali Madani, and Kevin~K Yang.
\newblock Flip: Benchmark tasks in fitness landscape inference for proteins.
\newblock {\em bioRxiv}, pages 2021--11, 2021.

\bibitem{notin2023proteinnpt}
Pascal Notin, Ruben Weitzman, Debora Marks, and Yarin Gal.
\newblock Proteinnpt: improving protein property prediction and design with non-parametric transformers.
\newblock {\em Advances in Neural Information Processing Systems}, 36, 2023.

\bibitem{nijkamp2023progen2}
Erik Nijkamp, Jeffrey~A Ruffolo, Eli~N Weinstein, Nikhil Naik, and Ali Madani.
\newblock Progen2: exploring the boundaries of protein language models.
\newblock {\em Cell systems}, 14(11):968--978, 2023.

\bibitem{yang2024convolutions}
Kevin~K Yang, Nicolo Fusi, and Alex~X Lu.
\newblock Convolutions are competitive with transformers for protein sequence pretraining.
\newblock {\em Cell Systems}, 15(3):286--294, 2024.

\bibitem{truong2023poet}
Timothy Truong~Jr and Tristan Bepler.
\newblock Poet: A generative model of protein families as sequences-of-sequences.
\newblock {\em Advances in Neural Information Processing Systems}, 36, 2023.

\bibitem{van2024fast}
Michel Van~Kempen, Stephanie~S Kim, Charlotte Tumescheit, Milot Mirdita, Jeongjae Lee, Cameron~LM Gilchrist, Johannes S{\"o}ding, and Martin Steinegger.
\newblock Fast and accurate protein structure search with foldseek.
\newblock {\em Nature Biotechnology}, 42(2):243--246, 2024.

\bibitem{fowler2014deep}
Douglas~M Fowler and Stanley Fields.
\newblock Deep mutational scanning: a new style of protein science.
\newblock {\em Nature methods}, 11(8):801--807, 2014.

\bibitem{rasmussen2010gaussian}
Carl~Edward Rasmussen and Hannes Nickisch.
\newblock Gaussian processes for machine learning (gpml) toolbox.
\newblock {\em The Journal of Machine Learning Research}, 11:3011--3015, 2010.

\bibitem{cao2019efficient}
Yueqi Cao, Didong Li, Huafei Sun, Amir~H Assadi, and Shiqiang Zhang.
\newblock Efficient curvature estimation for oriented point clouds.
\newblock {\em stat}, 1050:26, 2019.

\bibitem{sun2009concise}
Jian Sun, Maks Ovsjanikov, and Leonidas Guibas.
\newblock A concise and provably informative multi-scale signature based on heat diffusion.
\newblock In {\em Computer graphics forum}, volume~28, pages 1383--1392. Wiley Online Library, 2009.

\bibitem{devlin2018bert}
Jacob Devlin, Ming-Wei Chang, Kenton Lee, and Kristina Toutanova.
\newblock Bert: Pre-training of deep bidirectional transformers for language understanding.
\newblock {\em arXiv preprint arXiv:1810.04805}, 2018.

\bibitem{boucher2016quantifying}
Jeffrey~I Boucher, Daniel~NA Bolon, and Dan~S Tawfik.
\newblock Quantifying and understanding the fitness effects of protein mutations: Laboratory versus nature.
\newblock {\em Protein Science}, 25(7):1219--1226, 2016.

\bibitem{olson_comprehensive_2014}
C.~Anders Olson, Nicholas~C. Wu, and Ren Sun.
\newblock A comprehensive biophysical description of pairwise epistasis throughout an entire protein domain.
\newblock {\em Current Biology}, 24(22):2643--2651, November 2014.

\end{thebibliography}
\bibliographystyle{unsrt}


\newpage
\appendix

\begin{table*}[t]
    \centering
    \caption{\textbf{Average Spearman and Std. Error of Difference to Best Score.} 
     The best performing model is S3F-MSA, highlighted in \textbf{\first{red}}. $\Delta$ Spearman is the difference between the baseline models and the S3F-MSA. We followed the ProteinGym benchmarks to compute the std. error of difference to best score. In this table, it is the bootstrapped std. error of the differences between the Spearman of each model and the Spearman of S3F-MSA. 
    }
    \label{tab:proteingym_relative_s3f_msa}
    \begin{threeparttable}
    \begin{adjustbox}{max width=0.8\linewidth}
        \begin{tabular}{cccc}
            \toprule[2pt]
            \bf{Model} & \bf Avg. Spearman & \bf $\Delta$ Spearman & \bf Std. Err of Diff. to Best Score \\
            \midrule[1.5pt]
            ProGen2 XL &  0.391 & -0.105 & 0.010 \\
            CARP & 0.368 & -0.128 & 0.013 \\
            ESM2 & 0.414 & -0.082 & 0.013 \\
            \midrule
            ProteinMPNN &  0.258 & -0.238 & 0.013 \\
            MIF & 0.383 & -0.113 & 0.011 \\
            ESM-IF & 0.422 & -0.074 & 0.010 \\
            \midrule
            MIF-ST & 0.383 & -0.113 & 0.010 \\
            ProtSSN & 0.442 & -0.054 & 0.007 \\
            SaProt & 0.457 & -0.039 & 0.000 \\
            \midrule
            \bf \baseline & 0.454 & -0.042 & 0.005 \\
            \bf \model & 0.470 & -0.026 & 0.005 \\
            \midrule[1.5pt]
            DeepSequence & 0.419 & -0.077 & 0.007 \\
            EVE & 0.439 & -0.057 & 0.005 \\
            GEMME & 0.455 & -0.041 & 0.006 \\
            \midrule
            MSA Transformer & 0.434 & -0.062 & 0.011 \\
            Tranception L & 0.434 & -0.062 & 0.007 \\
            TranceptEVE L & 0.456 & -0.040 & 0.006 \\
            \midrule
            \bf \baseline-MSA & 0.487 & -0.009 & 0.001 \\
            \bf \modelmsa & \bf \first{0.496} & \first{0.000} & \first{0.000} \\
            \bottomrule[2pt]
        \end{tabular}
    \end{adjustbox}
    \end{threeparttable}
\end{table*}

\begin{figure}[b]
    \centering
    \includegraphics[width=0.8\linewidth]{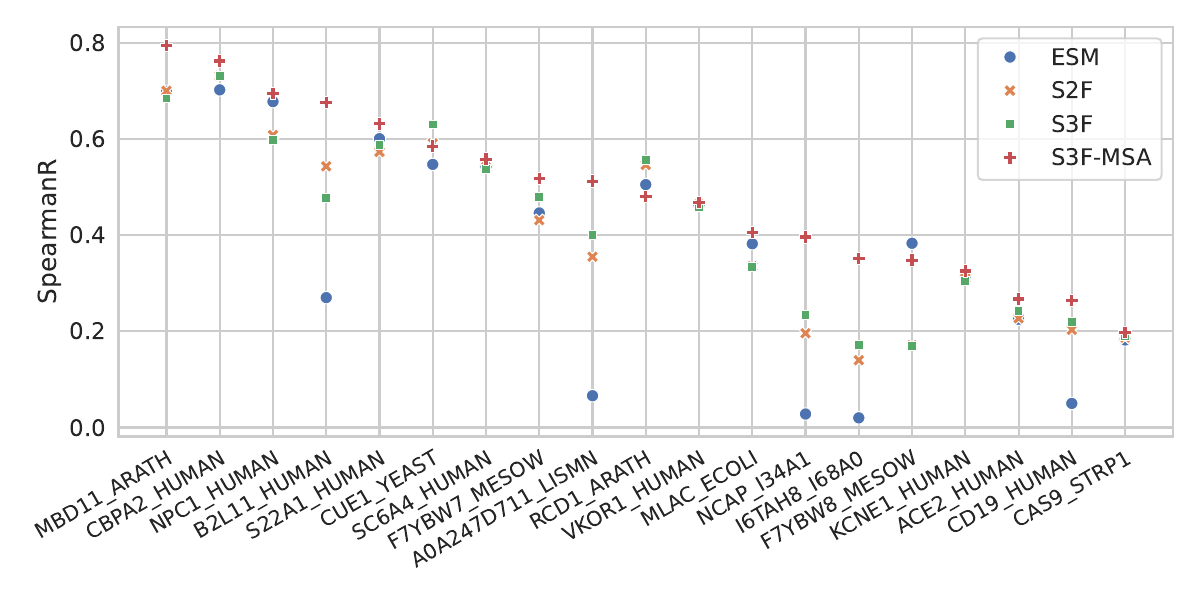}
    \caption{Spearmanr's rank correlation for ESM-2-650M, \baseline, \model and \modelmsa on 19 proteins with less than 30\% sequence similarity to the pre-training dataset.}
    \label{fig:ood_test}
\end{figure}

\section{Broader Impact}
\label{sec:impact}

The main objective of this project is to model protein fitness landscape more accurately with multi-scale representation learning.
Our approach utilizes structural information in the CATH dataset to build a multi-scale encoder with both sequence, structure and surface features.
This advantage allows for more comprehensive analysis of protein research and holds potential benefits for various real-world applications, including protein engineering, sequence and structure design.
It is important to acknowledge that powerful fitness prediction models can potentially be misused for harmful purposes, such as the design of dangerous drugs. We anticipate that future studies will address and mitigate these concerns.

\section{Additional Experimental Results}

\subsection{Performance on OOD test sets}

In Section~\ref{sec:generalization}, we select 23 out-of-distribution assays from ProteinGym by filtering with sequence similarity.
Here we report the performance on these assays in Figure~\ref{fig:ood_test}.
The results on assays with the same proteins are grouped together.

\subsection{Statistical Significance Analysis}

To quantify the statistical significance of the performance, we follow the same methodology as in ProteinGym and compute the non-parametric bootstrap standard error of the difference between the Spearman performance of a given model and that of the best overall model (10k bootstrap samples). Our performance delta with prior methods are all statistically significant, as shown in Table~\ref{tab:proteingym_relative_s3f_msa}.

\subsection{Ablation Study}

We provide ablation study results in Table~\ref{tab:ablation}, confirming the performance lift from the various modalities involved.
The ablation dropping both structure and surface features corresponds to ESM2 in Table~\ref{tab:proteingym}.
Note that surface message-passing is designed to capture fine-grained structural aspects that enhance the coarse-grained features learned by our S2F (sequence+structure) model. However, relying solely on these fine-grained features without the context from structural features, as we do in the ablation removing structural inputs, appears to be detrimental to performance.


\begin{table*}[h]
    \centering
    \caption{Ablation Study for \model.}
    \label{tab:ablation}
    \begin{adjustbox}{max width=0.6\linewidth}
        \begin{tabular}{lc}
            \toprule
            \bf Model & \bf Spearman \\
            \midrule
            \model w/o structure \& surface	& 0.414 \\
            \model w/o structure &	0.392 \\
            \model w/o surface & 0.454 \\
            \bf \model & \bf 0.470 \\
            \bottomrule
        \end{tabular}
    \end{adjustbox}
\end{table*}


\newpage
\section*{NeurIPS Paper Checklist}

\begin{enumerate}

\item {\bf Claims}
    \item[] Question: Do the main claims made in the abstract and introduction accurately reflect the paper's contributions and scope?
    \item[] Answer: \answerYes{} 
    \item[] Justification: The main claims about the experimental results of our methods are supported in Section~\ref{sec:exp}.

\item {\bf Limitations}
    \item[] Question: Does the paper discuss the limitations of the work performed by the authors?
    \item[] Answer: \answerYes{} 
    \item[] Justification: The limitations of this paper are discussed in Section~\ref{sec:conclusion}. 

\item {\bf Theory Assumptions and Proofs}
    \item[] Question: For each theoretical result, does the paper provide the full set of assumptions and a complete (and correct) proof?
    \item[] Answer: \answerNA{} 
    \item[] Justification: This paper does not include theoretical results. 

    \item {\bf Experimental Result Reproducibility}
    \item[] Question: Does the paper fully disclose all the information needed to reproduce the main experimental results of the paper to the extent that it affects the main claims and/or conclusions of the paper (regardless of whether the code and data are provided or not)?
    \item[] Answer: \answerYes{} 
    \item[] Justification: The implementation details are provided in Section~\ref{sec:method}. 

\item {\bf Open access to data and code}
    \item[] Question: Does the paper provide open access to the data and code, with sufficient instructions to faithfully reproduce the main experimental results, as described in supplemental material?
    \item[] Answer: \answerNo{} 
    \item[] Justification: The code, data and pre-trained model weights will be released upon acceptance. 

\item {\bf Experimental Setting/Details}
    \item[] Question: Does the paper specify all the training and test details (e.g., data splits, hyperparameters, how they were chosen, type of optimizer, etc.) necessary to understand the results?
    \item[] Answer: \answerYes{} 
    \item[] Justification: The evaluation setup are provided in Sections~\ref{sec:pretrain} and~\ref{sec:setup}.

\item {\bf Experiment Statistical Significance}
    \item[] Question: Does the paper report error bars suitably and correctly defined or other appropriate information about the statistical significance of the experiments?
    \item[] Answer: \answerNo{} 
    \item[] Justification: The statistical significance will be reported when the results are released on the ProteinGym benchmark. 

\item {\bf Experiments Compute Resources}
    \item[] Question: For each experiment, does the paper provide sufficient information on the computer resources (type of compute workers, memory, time of execution) needed to reproduce the experiments?
    \item[] Answer: \answerYes{} 
    \item[] Justification: The information about compute resources is provided in Section~\ref{sec:pretrain}.
    
\item {\bf Code Of Ethics}
    \item[] Question: Does the research conducted in the paper conform, in every respect, with the NeurIPS Code of Ethics \url{https://neurips.cc/public/EthicsGuidelines}?
    \item[] Answer: \answerYes{} 
    \item[] Justification: The research conform with the NeurIPS Code of Ethics.

\item {\bf Broader Impacts}
    \item[] Question: Does the paper discuss both potential positive societal impacts and negative societal impacts of the work performed?
    \item[] Answer: \answerYes{} 
    \item[] Justification: The broader impacts are discussed in Section~\ref{sec:impact}.

\item {\bf Safeguards}
    \item[] Question: Does the paper describe safeguards that have been put in place for responsible release of data or models that have a high risk for misuse (e.g., pretrained language models, image generators, or scraped datasets)?
    \item[] Answer: \answerNA{} 
    \item[] Justification: The paper poses no such risks. 

\item {\bf Licenses for existing assets}
    \item[] Question: Are the creators or original owners of assets (e.g., code, data, models), used in the paper, properly credited and are the license and terms of use explicitly mentioned and properly respected?
    \item[] Answer: \answerYes{} 
    \item[] Justification: The license of used datasets are provided. 

\item {\bf New Assets}
    \item[] Question: Are new assets introduced in the paper well documented and is the documentation provided alongside the assets?
    \item[] Answer: \answerNA{} 
    \item[] Justification: The paper does not release new assets. 

\item {\bf Crowdsourcing and Research with Human Subjects}
    \item[] Question: For crowdsourcing experiments and research with human subjects, does the paper include the full text of instructions given to participants and screenshots, if applicable, as well as details about compensation (if any)? 
    \item[] Answer: \answerNA{} 
    \item[] Justification: The paper does not involve crowdsourcing nor research with human subjects. 

\item {\bf Institutional Review Board (IRB) Approvals or Equivalent for Research with Human Subjects}
    \item[] Question: Does the paper describe potential risks incurred by study participants, whether such risks were disclosed to the subjects, and whether Institutional Review Board (IRB) approvals (or an equivalent approval/review based on the requirements of your country or institution) were obtained?
    \item[] Answer: \answerNo{} 
    \item[] Justification: The paper does not involve crowdsourcing nor research with human subjects. 

\end{enumerate}

\end{document}